\documentclass{article}
\usepackage{spconf,amsmath,graphicx}
\usepackage{afterpage}
\usepackage{csquotes}
\usepackage{algorithm}
\usepackage{algpseudocode}
\usepackage{color,soul}
\usepackage{multicol}
\usepackage{array}
\newcolumntype{P}[1]{>{\centering\arraybackslash}p{#1}}
\newcolumntype{M}[1]{>{\centering\arraybackslash}m{#1}}

\newcommand\blankpage{%
\null
\thispagestyle{empty}%
\addtocounter{page}{-1}%
\newpage}

\title{Predicting mechanical properties of Carbon Nanotube (CNT)
images Using Multi-Layer Synthetic Finite Element Model Simulations}
\name{\begin{tabular}{c}Kaveh Safavigerdini$^{1*}$\thanks{*Corresponding Author}, Koundinya Nouduri$^1$,  Ramakrishna Surya$^2$,  Andrew Reinhard$^2$, Zach Quinlan$^2$, \protect\\ Filiz Bunyak$^1$, Matthew R. Maschmann$^{2,3}$, Kannappan Palaniappan$^1$ \end{tabular}}
\address{
$^1$Department of Electrical Engineering and Computer Science\\
$^2$Department of Mechanical \& Aerospace Engineering\\
$^3$MU Materials Science and Engineering Institute\\
University of Missouri-Columbia\\
Columbia, Missouri, USA}
\blankpage
\begin{document}
%

\scalebox{2}{IEEE Copyright Notice}\\
\\
\\
~\copyright~2023 IEEE. Personal use of this material is permitted. Permission from IEEE must be obtained for all other uses, in any current or future media, including reprinting/republishing this material for advertising or promotional purposes, creating new collective works, for resale or redistribution to servers or lists, or reuse of any copyrighted component of this work in other works.
\\
\\
\textbf{Accepted to be published in}: the 2023 IEEE  International Conference on Image Processing (ICIP 2023); October  8-11, 2023; Kuala Lumpur, Malaysia;\\
https://2023.ieeeicip.org/
\maketitle
\begin{abstract}

We present a pipeline for predicting mechanical properties of vertically-oriented carbon nanotube (CNT) forest images using a deep learning model for artificial intelligence (AI)-based materials discovery. Our approach incorporates an innovative data augmentation technique that involves the use of multi-layer synthetic (MLS) or quasi-2.5D images which are generated by blending 2D synthetic images. The MLS images more closely resemble 3D synthetic and real scanning electron microscopy (SEM) images of CNTs but without the computational cost of performing expensive 3D simulations or experiments. Mechanical properties such as stiffness and buckling load for the MLS images are estimated using a physics-based model. The proposed deep learning architecture, CNTNeXt, builds upon our previous CNTNet neural network, using a ResNeXt feature representation followed by random forest regression estimator. Our machine learning approach for predicting CNT physical properties by utilizing a blended set of synthetic images is expected to outperform single synthetic image-based learning when it comes to predicting mechanical properties of real scanning electron microscopy images. This has the potential to accelerate understanding and control of CNT forest self-assembly for diverse applications.
                                
\end{abstract}
\section{Introduction}
\label{sec:intro}
The process-structure-property relationships governing carbon nanotube (CNT) forests remain poorly understood. In the previous work \cite{Hajilounezhad2021}, we used physics-based finite element model (FEM) simulations and machine learning methods to predict CNT forest mechanical stiffness and buckling load \cite{ferdosi2022axial} based on simulated CNT forest imagery.  The regression module of CNTNet \cite{Hajilounezhad2021} exhibited improved predictive performance for these mechanical properties, as indicated by a reduced root-mean-square error. While these methods produced promising results, the simulated imagery and mechanical data was for fictitious 2D CNT forests. The CNTNet produced poor predictive capability for scanning electron microscope (SEM) images of physical, 3D CNT forests.

\begin{figure}[H]
\centering 
\includegraphics[width=8.5cm]{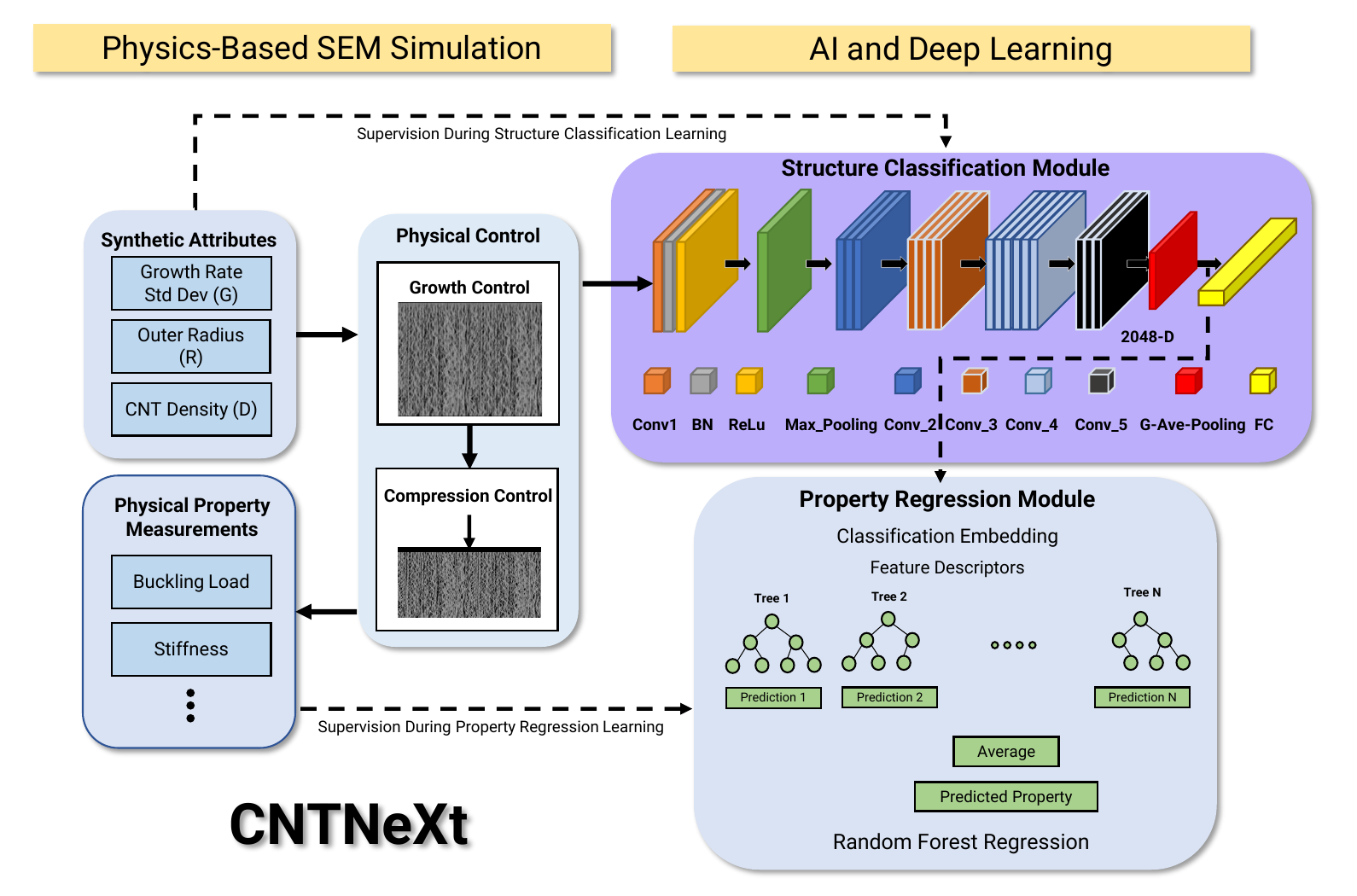}\\
\caption{Schematic of CNTNeXt modules}
\label{fig:CNTRNeXt}
\end{figure}

To increase the accuracy of CNTNet for predicting mechanical properties of SEM images, a novel combination of simulation images is presented here. To create quasi-images resembling SEM images from simulated CNT forests, various image processing techniques have been implemented. These techniques include blending with normalized Fibonacci sequence weights, color inversion, and applying Gaussian filters. The similarity between our multi-layer images and SEM images indicates that a deep learning model, such as our CNTNeXt model shown in Figure \ref{fig:CNTRNeXt}, has the potential to predict the mechanical properties of SEM images for high-throughput material discovery.

\begin{figure}[!ht]
\centering 
\includegraphics[width=8.5cm]{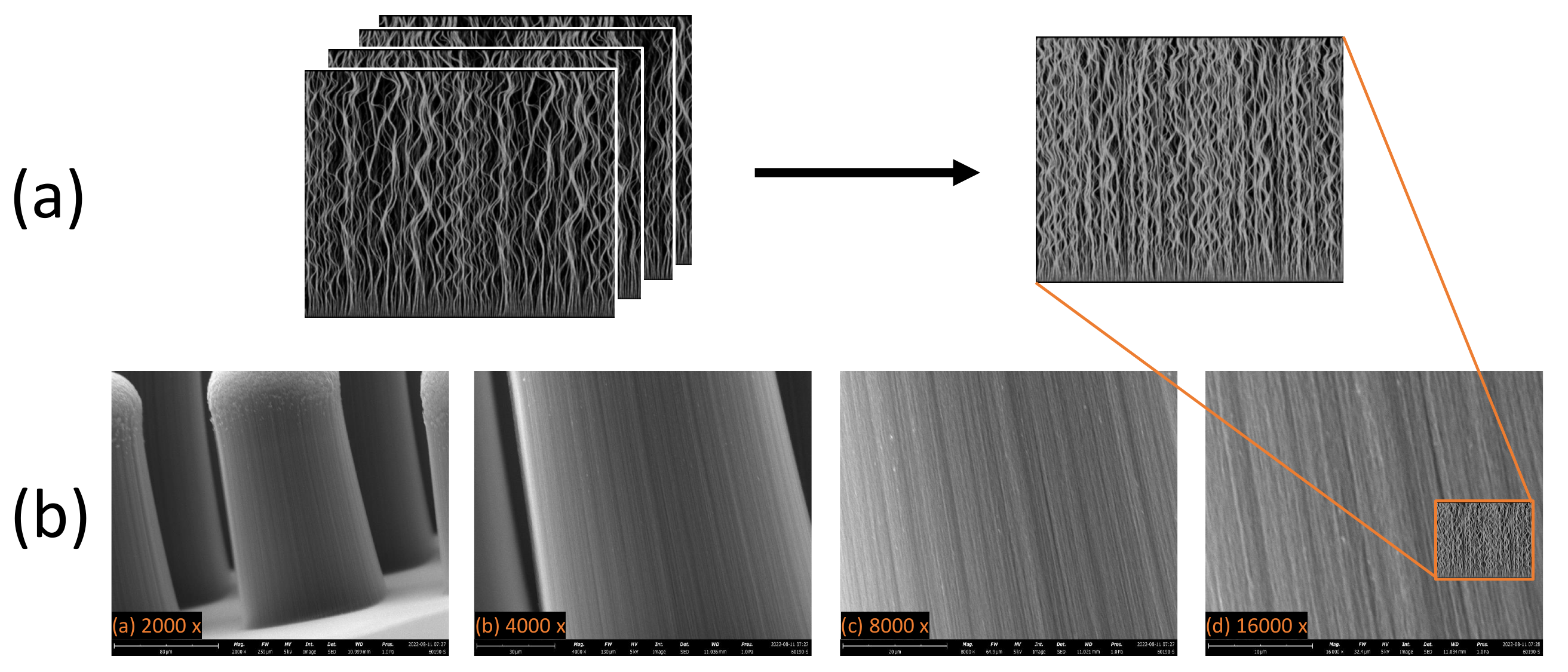}\\
\caption{Comparison between 2D synthetic, MLS and real SEM images. (a) MLS image created from four synthetic 2D images. (b) Desktop SEM (ThermoFisher Phenom Pharos) images of a CNT pillar at different zoom levels. Last image shows similarity between 16000x SEM and MLS images.}
\label{fig:SEM_Quasi_Similarity}
\end{figure}

Combining images to create multi-layer images is a technique used in many areas of research, including computer vision and image processing tasks \cite{Tian:21}. To the best of our knowledge, this is the first time that a multi-layer image dataset has been created that closely resembles 3D SEM images, as depicted in Figure \ref{fig:SEM_Quasi_Similarity} (a) and (b).

This method of combining simulated CNT images offers a unique approach to modeling and analyzing the growth of CNTs \cite{nguyen2023cnt}. The generated 2.5D images provide a valuable resource for testing and evaluating deep learning algorithms, as they closely resemble real SEM images of CNT growth. This approach can contribute to further understanding and improving the growth process of CNTs.

In this study, carbon nanotube forest simulation was conducted using a 2D mechanical finite element simulation with combinations of CNT growth rate distributions, CNT diameters, and CNT orientation angle inputs \cite{ maschmann2015integrated, hajilounezhad2019evaluating}. 2D CNT forest morphology images were generated after the final growth time step of the synthesis simulation. These images, representing one plane within a 3D CNT forest, were then combined to create 2.5D (quasi) images that resemble SEM images of real CNT forests. These 2.5D images serve as valuable candidates for deep learning testing and evaluation of real CNT images.

\section{Methodology}
\label{sec:Materials_Methods}

This section describes the creation of our novel multi-layer synthetic dataset, SEM images and CNTNeXt model.

\subsection{Multi Layer Synthetic Image dataset}
\label{ssec:multi_layer_dataset}
\begin{figure}[!ht]
\centering 
\includegraphics[width=8.5cm]{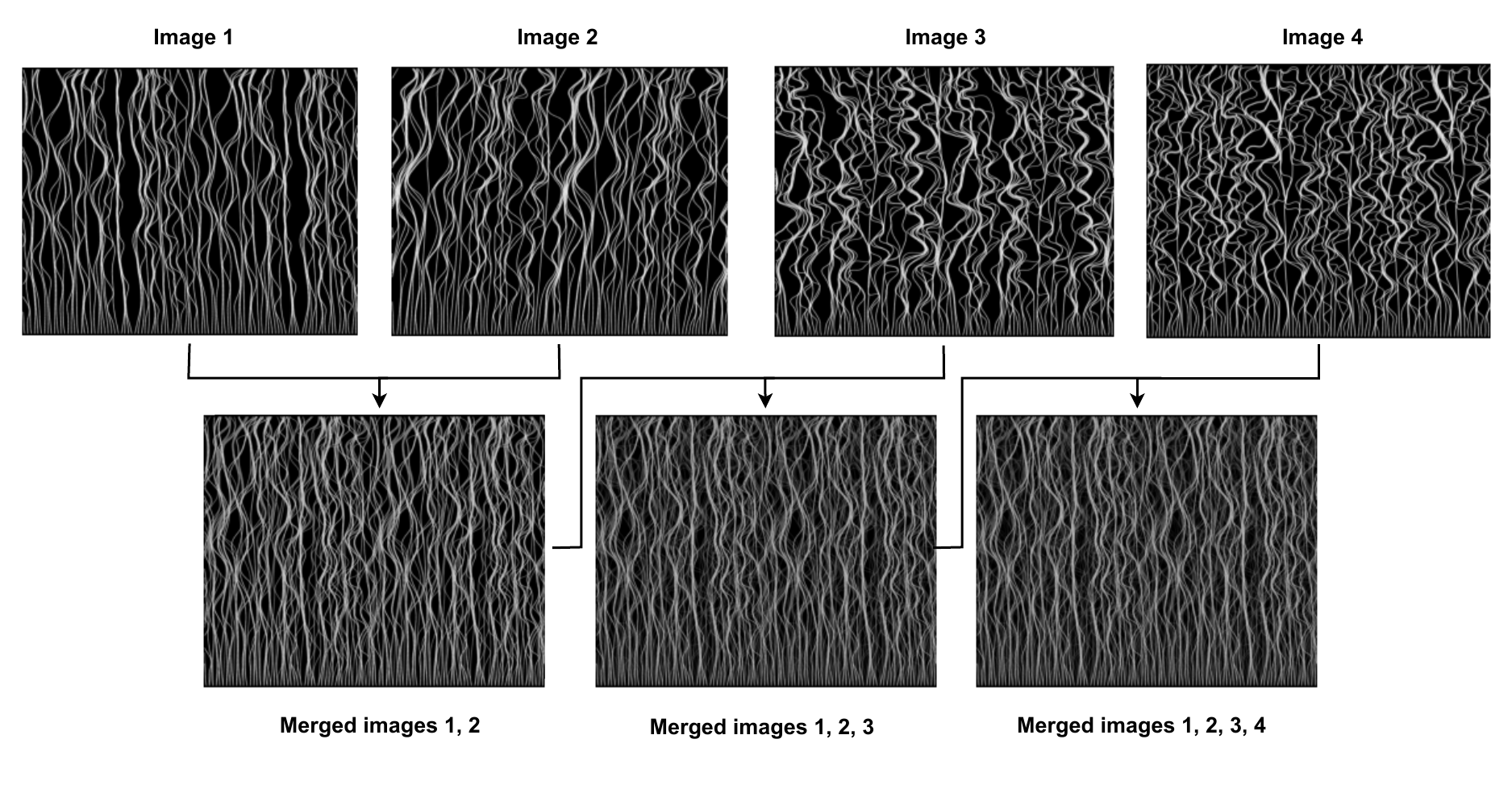}\\
\caption{Creating a multi-layer synthetic image or quasi-2.5D image by blending four 2D simulated images of CNTs using normalized Fibonacci weights.}
\label{fig:BlendingImgsProcess}
\end{figure}

\begin{figure}[H]
\centering 
\includegraphics[width=8cm]{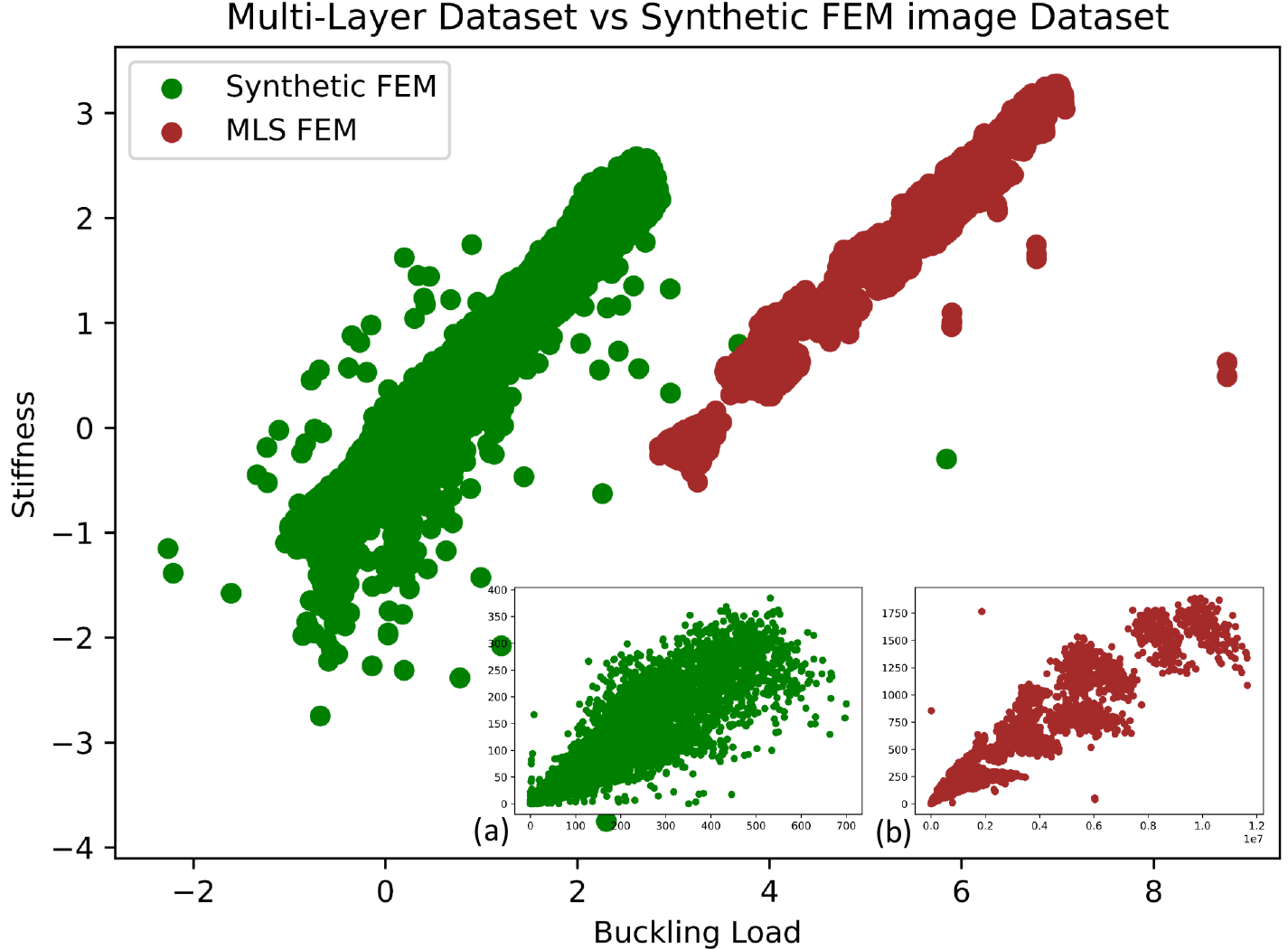}\\
\caption{Comparison of the Multi-Layer Image dataset and Synthetic FEM Image dataset using natural logarithmic scale. (a) Synthetic FEM dataset in normal scale, (b) Multi-Layer dataset in normal scale.}
\label{fig:synVsMLS}
\end{figure}

To create the MLS dataset, images from classes with identical CNT areal density, CNT radius, and growth rate distributions were selected for blending. The blending weight ratios were determined using a normalized Fibonacci sequence, resulting in visually appealing images compared to using equal weights. The blending process is demonstrated in Figure \ref{fig:BlendingImgsProcess}.
In order to calculate the equivalent buckling load and stiffness \cite{norouzi2020shape} of the MLS images, we employed the following formulas:
\begin{align}\label{eq:quasi_BuclingLoad}
    F_{eq} &= \big( \sum_{i=1}^{\mathcal{N}}{F_i} \big) *  \frac{\rho}{\mathcal{N}} * 100,\\
    S_{eq} &= \sum_{i=1}^{\mathcal{N}}{S_i},
\end{align}
where $F_{eq}$ represents the equivalent buckling load of the 2.5D stack of image layers, $\mathcal{N}$ is the number of layers, $F_i$ and $\rho$ represent the buckling load and density of layer $i$, respectively. Meanwhile, $S_{eq}$ stands for the equivalent stiffness and $S_i$ is the stiffness of layer $i$. Figure \ref{fig:synVsMLS} illustrates the comparison of mechanical properties between the 2.5D MLS and 2D Synthetic FEM Image datasets. The comparison is shown using both natural logarithmic and normal scales.

\begin{figure}[!ht]
\centering 
\includegraphics[width=8.5cm]{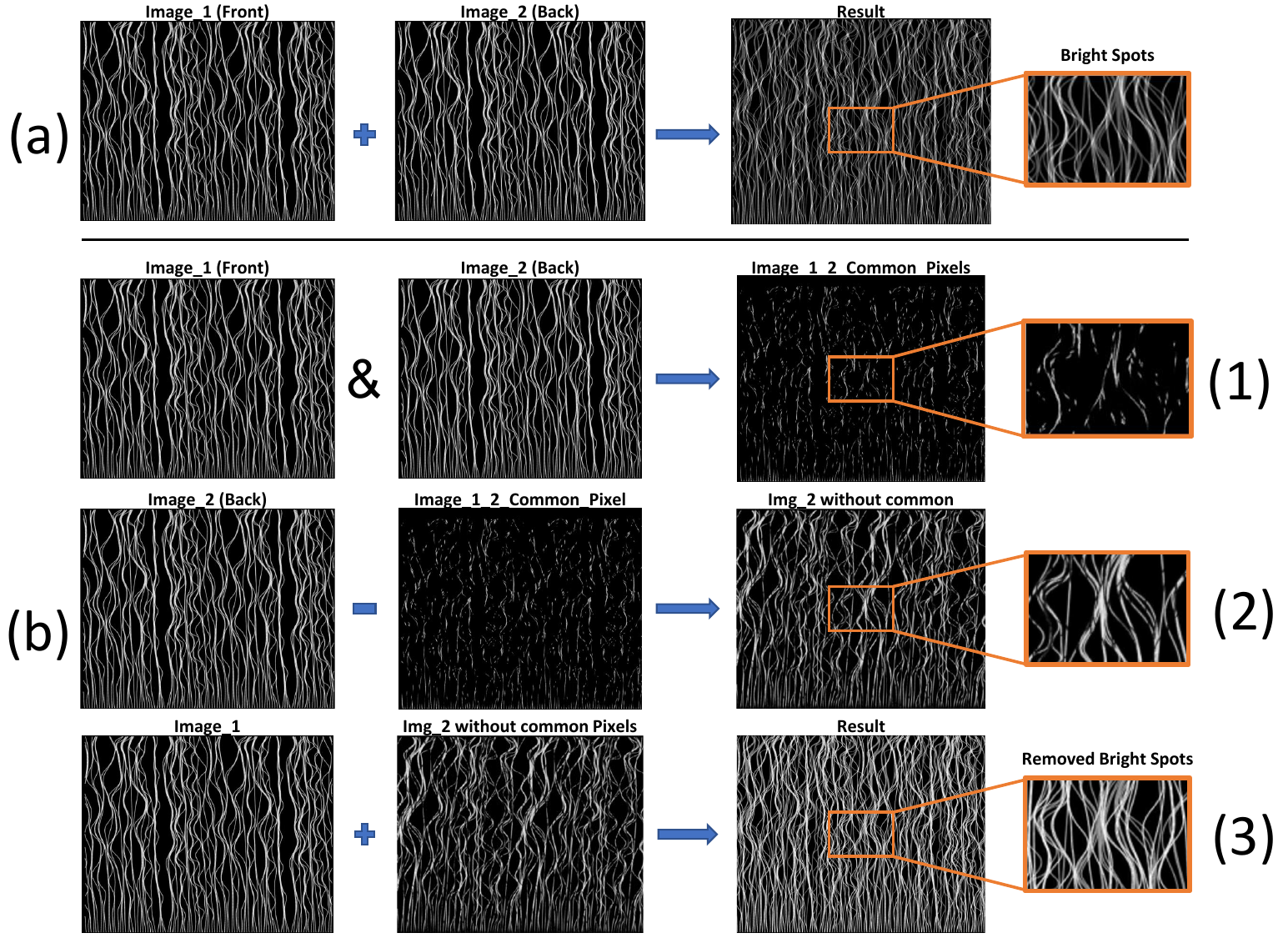}\\
\caption{Detecting and removing bright spots caused by overlapping images of 2D synthetic CNTs.}
\label{fig:TransProbSolve}
\end{figure}

When combining 2D synthetic carbon nanotube images using the OpenCV addWeight function, the overlapping pixels are summed up, resulting in the formation of high-intensity bright spots in the resulting 2.5D image, as depicted in Figure\ref{fig:TransProbSolve} (a). However, this phenomenon is not observed in real CNT forest images. In these images, the bright spots are correlated with a random particle distribution rather than overlapping CNTs.

Algorithm \ref{alg:tarnsProbSlove} is used to remove bright spot regions due to overlapping layer of 2D CNT images. Figure \ref{fig:TransProbSolve} (b), illustrates the process of resolving bright spots to generate more realistic and natural-looking MLS images. 
\begin{algorithm}[!ht]
\begin{small}
  \caption{Bright Spot Removal}
  \label{alg:tarnsProbSlove}
  \begin{enumerate}
    \itemsep0.1em
    \item Perform the OpenCV \textbf{bitwise AND} operation to identify common pixels between the two images. (Fig.~\ref{fig:TransProbSolve} (b)(1))
    \item Use the OpenCV \textbf{substract} function to remove the common pixels from the second image. (Fig.~\ref{fig:TransProbSolve} (b)(2))
    \item Utilize the OpenCV \textbf{addWeighted} function to blend the modified second image with the first image. (Fig.~\ref{fig:TransProbSolve} (b)(3))
  \end{enumerate}
\end{small}
\end{algorithm}
\subsection{Preparing SEM Images}
\label{ssec:Preparing_SEM_images}
We synthesized carbon nanotubes using different growth conditions, such as varying temperature, substrate type, and growth methods. We employed fixed and floating catalyst chemical vapor deposition to synthesize solid materials through gas-phase reactions, resulting in carbon nanotubes with different mechanical properties depending on the type of substrate and growth method used. We characterized the physical properties of the resulting carbon nanotubes by capturing images with a desktop SEM and conducting physical tests using a nanoindenter. Figure \ref{fig:SEM_Quasi_Similarity} (b) shows samples of SEM images at different magnifications.

\subsection{Deep Learning Model}
\label{ssec:outModel}
In our proposed approach, we use ResNeXt-50 both as a feature extractor and a classifier. ResNeXt is an enhanced version of ResNet \cite{he2016deep}, which introduces a new parameter called "cardinality" to achieve higher accuracy using fewer parameters than ResNet. Moreover, ResNeXt has demonstrated superior robustness to input data variations and can be effectively parallelized during training, making it a widely favored option for computer vision tasks \cite{xie2017aggregated}.



The suggested model employs Random Forest as a regression model due to its superior performance in dealing with high-dimensional data \cite{rf_tpami, kassim2016random}. 
Random Forest trains multiple decision trees in parallel, utilizing a combination of classification, bagging, and regression tree methods \cite{10.1007/978-3-030-89131-2_26, 7899614}.

\begin{figure*}[!ht]
  \includegraphics[width=\textwidth ]{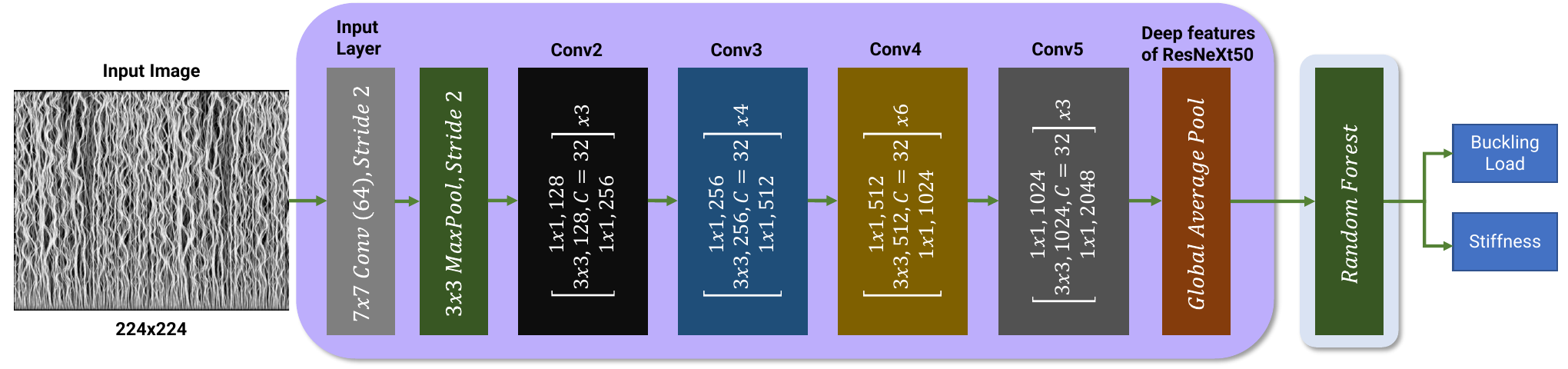}
  \caption{Proposed architecture incorporating deep features of ResNeXt-50 with a Random Forest regressor}
\label{fig:CNTRNeXtRF}
\end{figure*}

The proposed approach combines the strengths of deep learning and traditional machine learning techniques \cite{yaqoob2021resnet, Najafzadeh_2023_WACV}. Our approach involves extracting high-level features from the average pooling layer of the ResNeXt-50 model, which are derived from our multi-layer images. We then utilize a random forest regressor to predict buckling load and stiffness properties. The process is visually depicted in the figure \ref{fig:CNTRNeXtRF}. 

\section{Experimental Results}
\label{sec:Experimental_Results}
To train and test our pipeline, we followed a similar procedure to our previous work, CNTNet \cite{Hajilounezhad2021}. We partitioned our multi-layer image dataset into training $60\%$, validation $20 \%$, and testing $20 \%$ sets. Our MLS images were 907 × 725 pixels in size, and during training, we used random cropping augmentation to extract sub-images that were 224 × 224 pixels. For validation and testing, we performed center cropping of each image to make them 224 × 224 pixels. Several deep learning models were considered as feature extractors. Initially, we experimented VGG19 \cite{simonyan2014very}, and subsequently, we experimented ResNet50, ResNet101, and ResNeXt50 \cite{he2016deep, xie2017aggregated} as our model backbone \cite{saadabadi2023quality, kashiani2022robust}, for image texture-based categorization of CNT structures in the classification label assignment task. We used stochastic gradient optimization with the following hyperparameters: an initial learning rate of $10^{-3}$, weight decay of $10^{-4}$, and a momentum of 0.9 for gradient updating at each iteration. Our experimental results are presented in Table \ref{tbl:CNTNeXtClassificationResults}. 
\begin{table}[!ht]
\centering
\begin{tabular}{|c |c |c |} 
 \hline
Model &	Training (Acc \%)	&	Testing (Acc \%)	\\
 \hline
VGG19 \cite{simonyan2014very}     &    79.9    &    78.5    \\
 \hline
ResNet50 \cite{he2016deep}     &    83.4    &    83.1   \\
 \hline
ResNet101 \cite{he2016deep}    &    84.8    &    85.1  \\
 \hline
ResNeXt50 \cite{xie2017aggregated}    &    88.2    &    87.3    \\
 \hline
\end{tabular}
\caption{Classification result of different CNTNeXt's feature extractors on multi-layer image dataset}
\label{tbl:CNTNeXtClassificationResults}
\end{table}
The results show that the ResNeXt model had the highest classification accuracy, outperforming the other models. Therefore, we selected ResNeXt as the DL backbone for our proposed model. In order to evaluate the classification performance of CNTNeXt, we employed the standard overall accuracy metric defined as follows:

\begin{align}\label{eq:acc}
OA = \frac{\sum_{k=1}^{K}C_{k,k}}{\sum_{q=1}^{K}\sum_{k=1}^{K}C_{q,r}},
\end{align}
where $K$ is the number of classes, $C_{q,r}$ are the misclassification probabilities between the ground-truth class $q$ and the predicted class $r$, and the confusion matrix $C_{k,k}$ represent the correct classification probabilities \cite{Hajilounezhad2021}.

\begin{figure}[!ht]
\centering 
\includegraphics[width=6cm]{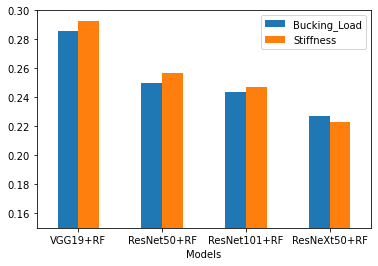}\\
\caption{RMSE of mechanical properties prediction of random forest with different deep architecture feature representation that were trained and tested on multi-layer synthetic images.}
\label{fig:CNTRNeXt_DiffModels}
\end{figure}


To train the random forest regressor decision trees, we used the 2048-dimension feature vector embedding from the last fully connected layer in the ResNeXt classification model. Our RF regression module in the CNTNeXt, builds an ensemble of regression trees with randomized feature selection to learn the quantitative relationship between elastic stiffness and buckling load values. Our RF ensemble included 1000 decision trees of varying depth, and we considered all 2048 features at each node split, with pure leaf nodes as the stopping condition during tree construction. The RF vector regressor is trained to predict two properties, buckling load and stiffness. 

We evaluated the performance of the RF regressor with the feature extractors mentioned above, and the results are shown in Figure \ref{fig:CNTRNeXt_DiffModels}.
Our CNTNeXt model, a combination of RF and ResNeXt, demonstrates improved performance in terms of RMSE values \cite{toxics10120746} for stiffness and buckling load. The RMSE metric is defined in equation \ref{eq:rmse}, where n represents the total number of samples, $y_{i}$ denotes the ground truth, and $x_{i}$ signifies the prediction for the $i^{th}$ sample.




\begin{align}\label{eq:rmse}
    RMSE = \sqrt{(\frac{1}{n})\sum_{i=1}^{n}(y_{i} - x_{i})^{2}}
\end{align}

\section{Conclusions}
\label{sec:Conclusion}
In this study, we have presented a pipeline for predicting mechanical properties of vertically-oriented carbon nanotube (CNT) forest images using a deep learning model. Our approach leverages a novel data augmentation method with multi-layer synthetic images, which more closely resemble 3D synthetic and real scanning electron microscopy images of CNTs but without the computational cost of performing expensive 3D simulations or experiments. 

We also proposed a deep learning architecture, CNTNeXt, which builds upon our previous CNTNet neural network, and incorporates a ResNeXt feature representation followed by random forest regression estimator. Our results demonstrate that the ResNeXt model has the highest classification accuracy, outperforming other models, and combining with RF leads to lower RMSE values for both stiffness and buckling load. This suggests that our CNTNeXt model can effectively predict the mechanical properties of real SEM images of CNTs, and has the potential to accelerate understanding and control of CNT forest self-assembly for diverse applications.


\bibliographystyle{IEEEbib}

\bibliography{ref}

\end{document}